\documentclass{article}

      \usepackage[preprint,nonatbib]{neurips_2019}
\pdfoutput=1

\usepackage[
    backend=biber,
    style=numeric,
    sorting=none,
    sortlocale=de_DE,
    natbib=true,
    url=false,
    doi=true,
    eprint=false
]{biblatex}
\usepackage[utf8]{inputenc} 
\usepackage[T1]{fontenc}    
\usepackage{hyperref}       
\usepackage{url}            
\usepackage{booktabs}       
\usepackage{amsfonts}       
\usepackage{nicefrac}       
\usepackage{microtype}      

\usepackage{graphicx}
\usepackage{mathtools}
\DeclarePairedDelimiter{\abs}{\lvert}{\rvert}
\usepackage{subcaption}
\newcommand{\R}{\mathbb{R}}
\DeclareMathOperator{\sign}{sign}
\usepackage[backgroundcolor=red]{todonotes}   
\usepackage{authblk}

\title{Parameterized Structured Pruning for Deep Neural Networks}

\author[1]{G\"unther Schindler}
\author[2]{Wolfgang Roth}
\author[2]{Franz Pernkopf}
\author[1]{Holger Fr\"oning}
\affil[1]{Institute of Computer Engineering,
					Ruprecht Karls University, Heidelberg, Germany}
\affil[2]{Signal Processing and Speech Communication Laboratory,
					Graz University of Technology, Austria}

\begin{document}

\maketitle

\begin{abstract}
As a result of the growing size of Deep Neural Networks (DNNs), the gap to hardware capabilities in terms of memory and compute increases.
To effectively compress DNNs, quantization and connection pruning are usually considered.
However, unconstrained pruning usually leads to unstructured parallelism, which maps poorly to massively parallel processors, and substantially reduces the efficiency of general-purpose processors.
Similar applies to quantization, which often requires dedicated hardware.

We propose Parameterized Structured Pruning (PSP), a novel method to dynamically learn the shape of DNNs through structured sparsity.
PSP parameterizes structures (e.g. channel- or layer-wise) in a weight tensor and leverages weight decay to learn a clear distinction between important and unimportant structures.
As a result, PSP maintains prediction performance, creates a substantial amount of sparsity that is structured and, thus, easy and efficient to map to a variety of massively parallel processors, which are mandatory for utmost compute power and energy efficiency.
PSP is experimentally validated on the popular CIFAR10/100 and ILSVRC2012 datasets using ResNet and DenseNet architectures, respectively.
\end{abstract}


\section{Introduction}
Deep Neural Networks (DNNs) are widely used for many applications including object recognition \cite{Krizhevsky:2012:ICD:2999134.2999257}, speech recognition \cite{hinton12} and robotics \cite{lenz2016deep}. 
An empirical observation is that DNNs, trained by Stochastic Gradient Descent (SGD) from random initialization, are remarkable successful in fitting training data \cite{DBLP:journals/corr/ZhangBHRV16}.
The ability of  modern DNNs to excellently fit training data is suspected to be due to heavy over-parameterization, i.e., using more parameters than the total number of training samples, since there always exists parameter choices that achieve a training error of zero \cite{DBLP:journals/corr/abs-1811-03962}.
In particular, Li et al. \cite{DBLP:journals/corr/abs-1808-01204} showed that SGD finds nearly-global optimal solution on the training data, as long as the network is over-parameterized which can be extended to test data as well.

While over-parameterization is essential for the learning ability of neural networks, it results in extreme memory and compute requirements for training (development) as well as inference (deployment).
Recent research showed, e.g. \cite{DBLP:journals/corr/abs-1811-06992}, that training can be scaled to up to 1024 accelerators operating in parallel, resulting in a development phase not exceeding a couple of minutes, even for large-scale image classification.
However, the deployment has usually much harder constraints than the development, as energy, space and monetary cost are scarce in mobile devices.

Model compression methods are targeting this issue by training an over-parameterized model and compressing it for deployment.
Popular compression methods are pruning \cite{DBLP:journals/corr/HanPTD15, DBLP:journals/corr/GuoYC16}, quantization \cite{DBLP:journals/corr/abs-1807-10029}, knowledge distillation \cite{HinVin15Distilling}, and low-rank factorization \cite{DBLP:journals/corr/JaderbergVZ14, DBLP:journals/corr/DentonZBLF14}, with the first two being most popular due to their extreme efficiency. Pruning connections \cite{DBLP:journals/corr/HanPTD15, DBLP:journals/corr/GuoYC16} achieves impressive theoretical compression rates through fine-grained sparsity (Fig.~\ref{fig:sub1}) without sacrificing prediction performance, but has several practical drawbacks such as indexing overhead, load imbalance and random memory accesses:
(i) Compression rates are typically reported without considering the space requirement of additional data structures to represent non-zero weights. For instance, using indices, a model with 8-bit weights, 8-bit indices and 75\% sparsity saves only 50\% of the space, while a model with 50\% sparsity does not save memory at all.
(ii) It is a well-known problem that massively parallel processors show notoriously poor performance when the load is not well balanced. Unfortunately, since the end of Dennard CMOS scaling \cite{1050511}, massive parallelization is mandatory for a continued performance scaling.
(iii) Sparse models increase the amount of randomness in memory access patterns, preventing caching methods which rely on predictable strides from being effective. As a result, the amount of cache misses increases the average memory access latency and energy consumption, as off-chip accesses are 10-100 time higher in terms of latency, respectively 100-1000 times higher in terms of energy consumption \cite{6757323}. Quantization has recently received plenty of attention and reduces computational complexity as additions scale linearly and multiplications scale quadratically with the number of bits \cite{SzeCYE17}. However, in comparison, pruning avoids a computation completely.

Structured pruning methods can prevent these drawbacks by inducing sparsity in a hardware-friendly way: Fig. \ref{fig:sub2}-\ref{fig:sub5} illustrate exemplary a 4-dimensional convolution tensor (see \cite{DBLP:journals/corr/ChetlurWVCTCS14} for details on convolution lowering), where hardware-friendly sparsity structures are shown as channels, layers, etc.
However, pruning whole structures in a neural network is not as trivial as pruning individual connections and usually causes high accuracy degradation under mediocre compression constraints.
Structured pruning methods can be roughly clustered into two categories: re-training-based and regularization-based methods (see Sec.~\ref{sec:related_work} for details).
Re-training-based methods aim to remove structures by minimizing the pruning error in terms of changes in weight, activation, or loss, respectively, between the pruned and the pre-trained model.
Regularization-based methods train a randomly initialized model and apply regularization, usually an $\ell_1$ penalty, in order to force structures to zero. 
This work introduces a new regularization-based method based on learned parameters for structured sparsity without substantial increase in training time.
Our approach differs from previous methods, as we explicitly parameterize certain structures of weight tensors and regularize them with weight decay, enabling a clear distinction between important and unimportant structures.
Combined with threshold-based magnitude pruning and a straight-through gradient estimator (STE) \cite{DBLP:journals/corr/BengioLC13}, we can remove a substantial amount of structure while maintaining the classification accuracy.
We evaluate the proposed method based on state-of-the-art Convolutional Neural Networks (CNNs) like ResNet \cite{DBLP:journals/corr/HeZRS15} and DenseNet \cite{DBLP:journals/corr/HuangLW16a}, and popular datasets like CIFAR-10/100 \cite{cifar} and ILSVRC2012 \citep{DBLP:journals/corr/RussakovskyDSKSMHKKBBF14}.

The remainder of this work is structured as follows:
Related work is summarized in Sec.~\ref{sec:related_work}.
In Sec.~\ref{sec:param_pruning} we introduce the parameterization and regularization approach together with the pruning method.
We present experimental results in Sec.~\ref{sec:experiments}, before we conclude in Sec.~\ref{sec:conclusion}.



\section{Related Work}
\label{sec:related_work}

\textbf{Re-training-based methods:}
In \cite{DBLP:journals/corr/HuPTT16} it is proposed to prune neurons based on their average percentage of zeros activations.
Li et al. \cite{DBLP:journals/corr/LiKDSG16} evaluate the importance of filters by calculating its absolute weight sum.
Mao et al. \cite{DBLP:journals/corr/MaoHPLLWD17} prune structures with the lowest $\ell_1$ norm.
Channel Pruning (CP) \cite{DBLP:journals/corr/HeZS17} uses an iterative two-step algorithm to prune each layer by a LASSO regression based channel selection and least square reconstruction.
Structured Probabilistic Pruning (SPP) \cite{DBLP:journals/corr/abs-1709-06994} introduces a pruning probability for each weight where pruning is guided by sampling from the pruning probabilities.
Soft Filter Pruning (SFP) \cite{DBLP:journals/corr/abs-1808-06866} enables pruned filters to be updated when training the model after pruning, which results in larger model capacity and less dependency on the pre-trained model.
Layer-Compensated Pruning (LCP) \cite{DBLP:journals/corr/abs-1810-00518} leverages meta-learning to learn a set of latent variables that compensate for approximation errors.
ThiNet \cite{DBLP:journals/corr/LuoWL17} shows that pruning filters based on statistical information calculated from the following layer is more accurate than using statistics of the current layer.
Discrimination-aware Channel Pruning (DCP) \cite{DBLP:journals/corr/abs-1810-11809} selects channels based on their discriminative power.

\textbf{Regularization-based methods:}
Group LASSO \cite{Yuan06modelselection} allows predefined groups in a model to be selected together. Adding an $\ell_1$ penalty to each group is a heavily used approach for inducing structured sparsity in CNNs \cite{DBLP:journals/corr/LebedevL15,DBLP:journals/corr/WenWWCL16,Jenatton:2011:SVS:1953048.2078194,NIPS2016_6372}.
Network Slimming \cite{DBLP:journals/corr/abs-1708-06519}, Huang et al. \cite{DBLP:journals/corr/HuangW17aa} and MorphNet \cite{DBLP:journals/corr/abs-1711-06798} apply $\ell_1$ regularization on coefficients of batch-normalization layers in order to create sparsity in a structured way.



\section{Parameterized Pruning}
\label{sec:param_pruning}
DNNs are constructed by layers of stacked processing units, where each unit computes an activation function of the form
\begin{equation}
z = g(\textbf{W} \oplus \textbf{x}) ,
\end{equation}
where $\textbf{W}$ is a weight tensor, $\textbf{x}$ is an input tensor, $\oplus$ denotes a linear operation, e.g., a convolution, and $g(\cdot)$ is a non-linear function.
Modern neural networks have very large numbers of these stacked compute units, resulting in huge memory requirements for the weight tensors $\textbf{W}$ and compute requirements for the linear operations $\textbf{W} \oplus \textbf{x}$.
In this work, we aim to learn a structured sparse substitute $\textbf{Q}$ for the weight tensor $\textbf{W}$, so that there is only minimal overhead for representing the sparsity pattern in $\textbf{Q}$ while retaining computational efficiency using dense tensor operations.
For instance, by setting all weights at certain indices of the tensor to zero, it suffices to store the indices of non-zero elements only once for the entire tensor $\textbf{Q}$ an not for each individual dimension separately.
By setting all weights connected to an input feature map to zero, the corresponding feature map can effectively be removed without the need to store any indices at all.

\subsection{Hardware-friendly structures in CNNs}
We consider CNNs with $R \times S$ filter kernels, $C$ input and $K$ output feature maps.
Different granularities of structured sparsity yield different flexibilities when mapped to hardware.
In this work, we consider only coarse-grained structures such as column, channel and layer pruning, that can be implemented using off-the-shelf libraries on general-purpose hardware or shape pruning for direct convolutions on re-configurable hardware.

Convolutions are usually lowered onto matrix multiplication in order to explore data locality and the massive amounts of parallelism in general-purpose GPUs or specialized processors like TPUs \cite{DBLP:journals/corr/JouppiYPPABBBBB17}.
The reader may refer to the work of Chetlur et al. \cite{DBLP:journals/corr/ChetlurWVCTCS14} for a detailed explanation.
Consequently, the computation of all structured sparsities that are explored in this work can be lowered to dense matrix multiplications.

\begin{figure*}
\centering
\begin{subfigure}{.18\textwidth}
  \centering
  \includegraphics[width=0.8\linewidth]{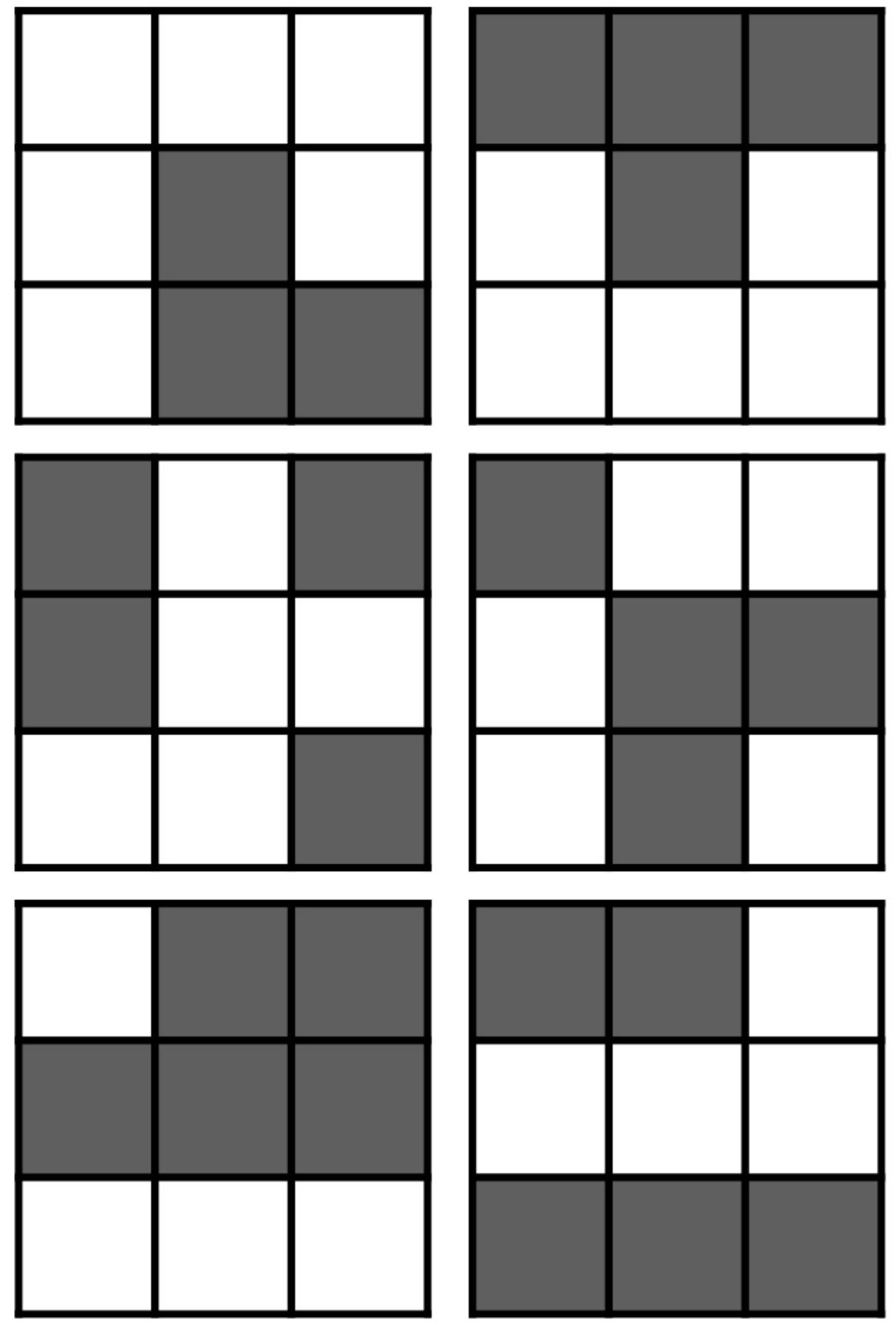}
  \caption{Weight pruning}
  \label{fig:sub1}
\end{subfigure}%
\begin{subfigure}{.18\textwidth}
  \centering
  \includegraphics[width=0.8\linewidth]{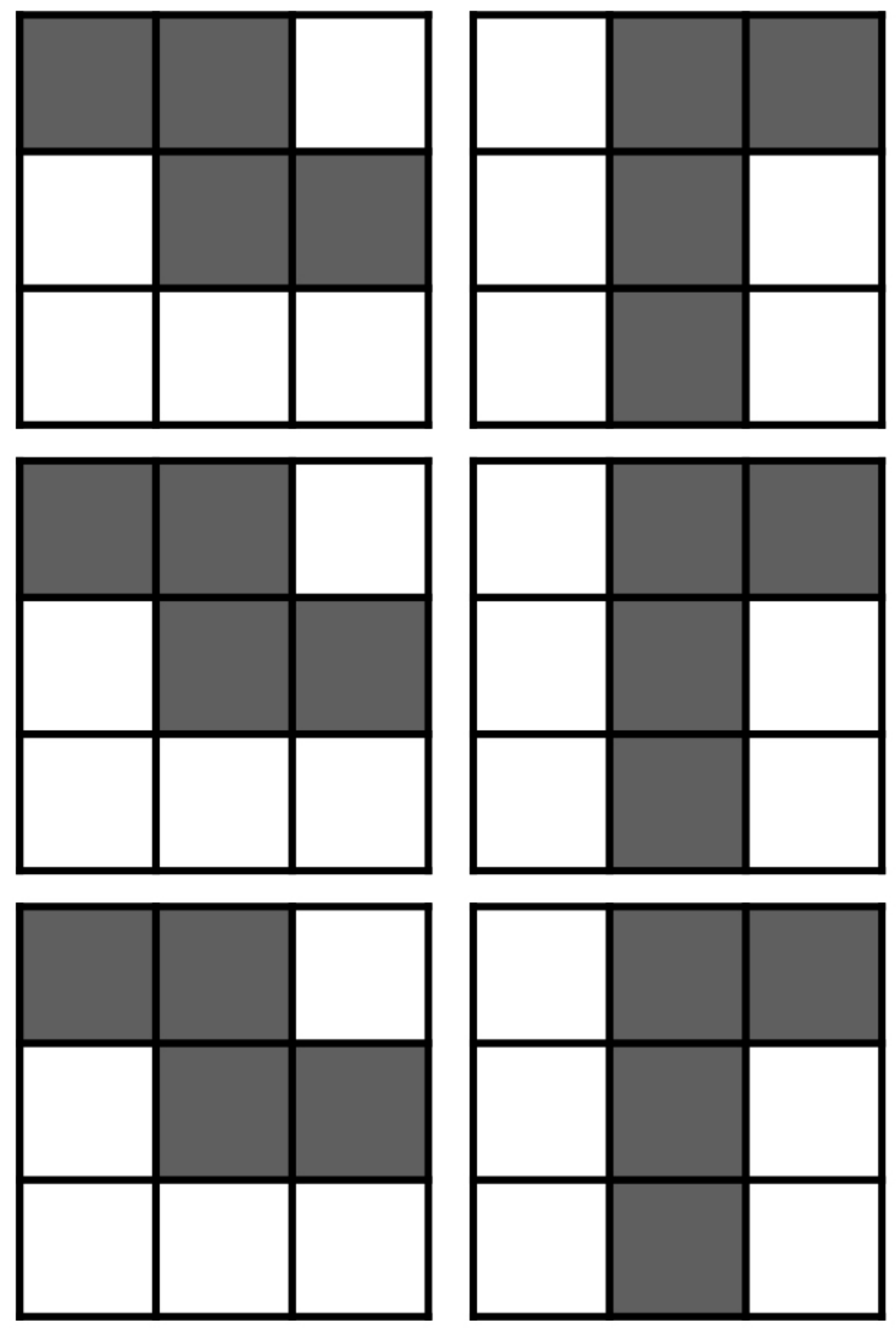}
  \caption{Column pruning}
  \label{fig:sub2}
 \end{subfigure}%
\begin{subfigure}{.18\textwidth}
  \centering
  \includegraphics[width=0.8\linewidth]{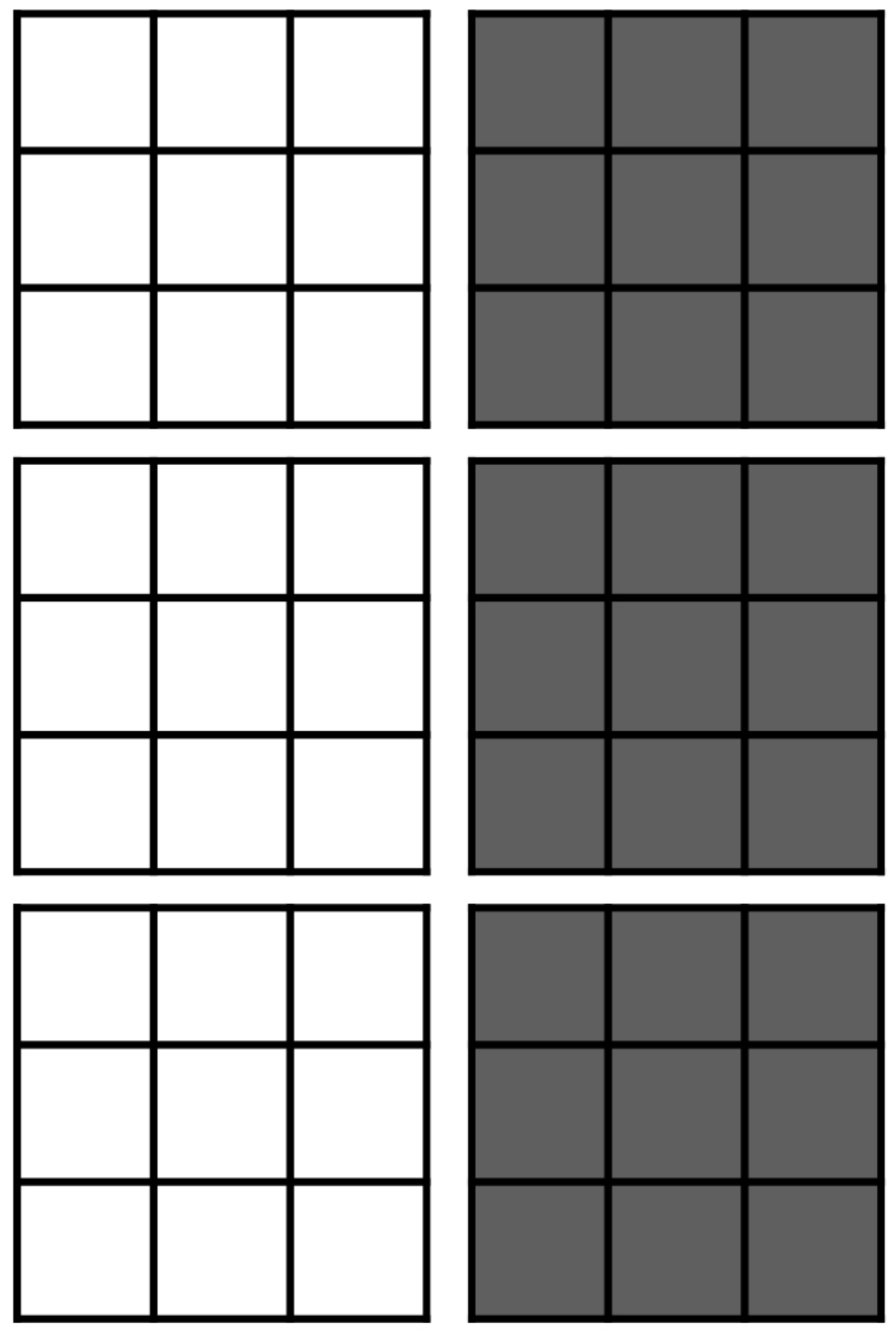}
  \caption{Channel pruning}
  \label{fig:sub3}
\end{subfigure}%
\begin{subfigure}{.18\textwidth}
  \centering
  \includegraphics[width=0.8\linewidth]{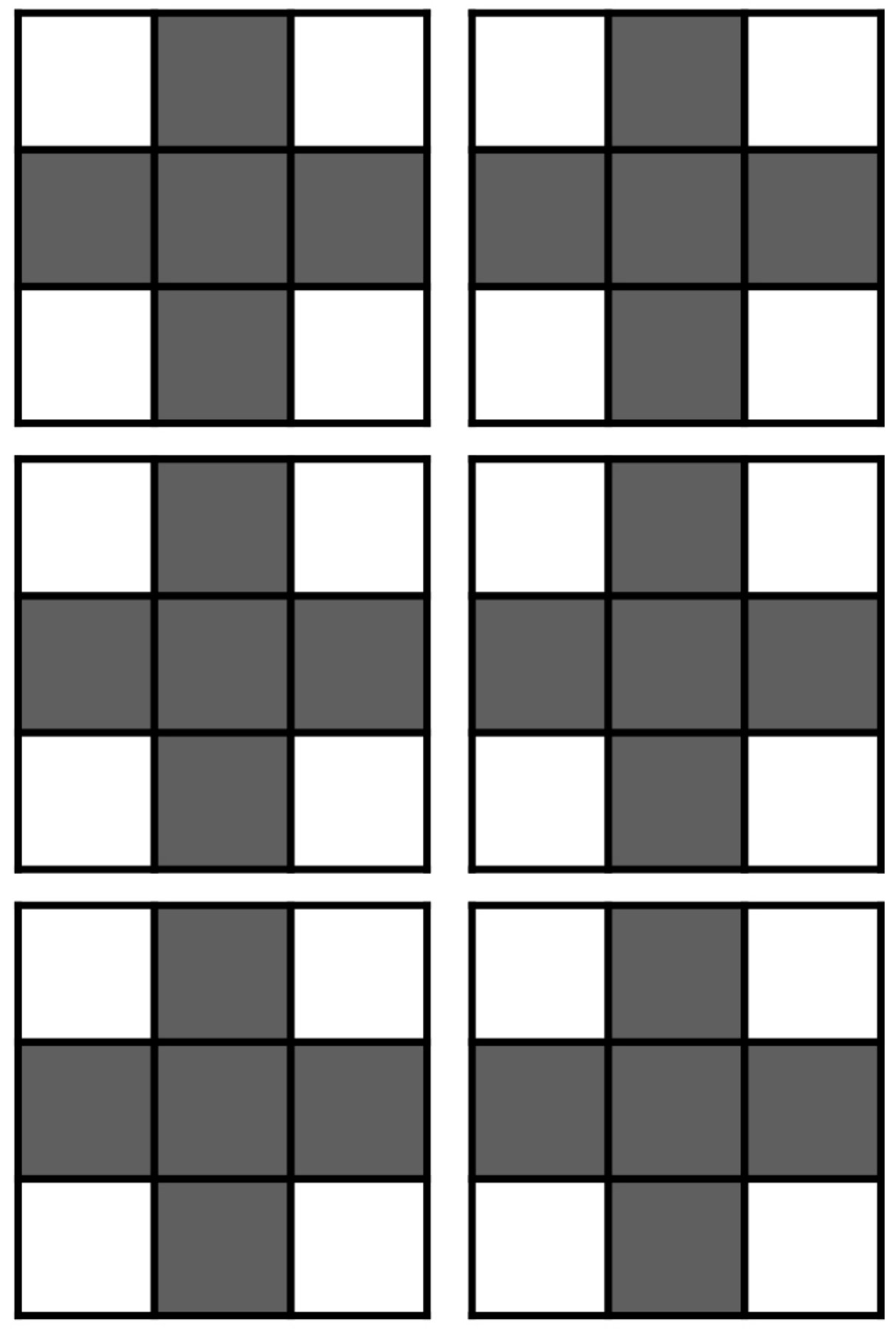}
  \caption{Shape pruning}
  \label{fig:sub4}
\end{subfigure}
\begin{subfigure}{.18\textwidth}
  \centering
  \includegraphics[width=0.8\linewidth]{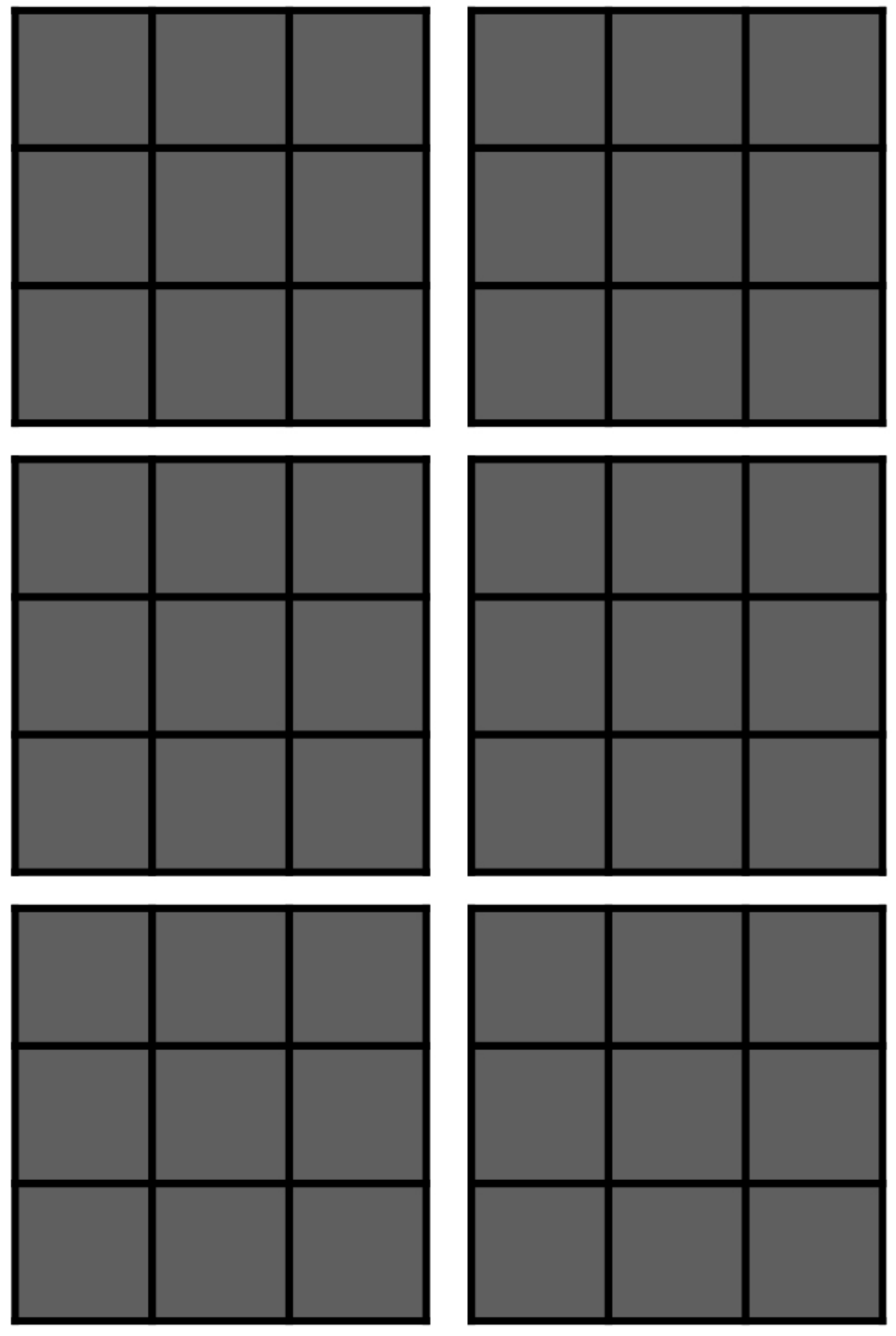}
  \caption{Layer pruning}
  \label{fig:sub5}
\end{subfigure}
\caption{Illustration of fine-grained (Fig. \ref{fig:sub1}) and several structured forms of sparsity (Fig. \ref{fig:sub2}-\ref{fig:sub4}) for a 4-dimensional convolution tensor. The large squares represent the kernels, and the corresponding horizontal and vertical dimensions represent the number of input feature and output feature maps, respectively. The computation of all structured forms of sparsity can be lowered to matrix multiplications (independent of stride and padding).}
\label{fig:distributions}
\end{figure*}

Column pruning refers to pruning weight tensors in a way that a whole column of the flattened weight tensor and the respective row of the input data can be removed (Fig. \ref{fig:sub2}).
Channel pruning refers to removing a whole channel in the weight tensor and the respective input feature map (Fig. \ref{fig:sub3}).
Shape pruning targets to sparsify filter kernels per layer equally (Fig. \ref{fig:sub4}), which can be mapped onto re-configurable hardware.
Layer pruning simply removes whole layers of a DNN (Fig. \ref{fig:sub5}).
The proposed approach is not restricted to these forms of sparsity, arbitrary structures and combinations of different structures are possible.
Other structured sparsites, but more fine-grained, are explored by Mao et al. \cite{DBLP:journals/corr/MaoHPLLWD17}.

\subsection{Parameterization}
Identifying the importance of certain structures in neural networks is vital for the prediction performance of structured-pruning methods.
Our approach is to train the importance of structures by parameterizing and optimizing them together with the weights using backpropagation.
Therefore, we divide the tensor $\mathbf{W}$ into subtensors $\{\mathbf{w}_i\}$ so that each $\mathbf{w}_i = (w_{i,j})_{j=1}^m$ constitutes the $m$ weights of structure $i$.
During forward propagation, we substitute $\textbf{w}_i$ by the structured sparse tensor $\textbf{q}_i$ as
\begin{equation}
 \textbf{q}_i = \textbf{w}_i \nu_i
\end{equation}
with
\begin{equation}
\label{eq:threshold}
\nu_i(\alpha_i) = \begin{cases}
    0        \quad & \abs{\alpha_i} < \epsilon 	\\
    \alpha_i \quad & \abs{\alpha_i} \geq \epsilon \\
\end{cases},
\end{equation}
where $\alpha_i$ is the \emph{dense structure parameter} associated with structure $i$ and $\epsilon$ is a tuneable pruning threshold.
As the threshold function $\nu_i$ is not differentiable at $\pm \epsilon$ and the gradient is zero in $[-\epsilon, \epsilon]$,
we approximate the gradient of $\nu_i$ by defining a STE as
\begin{equation}
\frac{\partial E}{\partial \nu_i} = \frac{\partial E}{\partial \alpha_i}.
\end{equation}
We use the sparse parameters $\nu_i$ for forward and backward propagation and update the respective dense parameters $\alpha_i$ based on the gradients of $\nu_i$.
Updating the dense structure parameters $\alpha_i$ instead of the sparse parameters $\nu_i$ is beneficial because improperly pruned structures can reappear if $\alpha_i$ moves out of the pruning interval $[-\epsilon, \epsilon]$, resulting in faster convergence to a better performance.
Following the chain rule, the gradient of the dense structure parameters $\alpha_i$ is:
\begin{equation}
\label{eq:conv0}
\frac{\partial E}{\partial \alpha_i} = \sum_{j=1}^{m} \frac{\partial E}{\partial w_{i,j}} \mbox{   },
\end{equation}
where $E$ represents the objective function.
As a result, the dense structure parameters $\alpha_i$ descent towards the predominant direction of the structure weights.
Training the structures introduces additional parameters, however, during inference they are folded into the weight tensors, resulting in no extra memory or compute costs.
The dense structure parameters for individual structures and their corresponding gradients are shown in Table~\ref{tab:gradient}. 
Note that layer pruning is only applicable to multi-branch neural network architectures.
\begin{table}[h!]
 \centering
 \scriptsize
 \caption{Representation of the dense structure parameters and the gradient calculation.}
 \label{tab:gradient}
 \begin{tabular}{|c|c|c|}
  \hline Pruning method & Dense structure parameter & Gradient \\
  \hline \hline Column pruning & $\mathbf{\alpha} \in \R^{R \times S \times C}$ & $\partial E / \partial \alpha_{r,s,c} = \sum_{k=1}^{K} \partial E / \partial W_{k,c,r,s}$ \\
  \hline Channel pruning & $\mathbf{\alpha} \in \R^{C}$ & $\partial E / \partial \alpha_{c} = \sum_{k=1}^{K} \sum_{r=1}^{R} \sum_{s=1}^{S} \partial E / \partial W_{k,c,r,s}$ \\
  \hline Shape pruning & $\alpha \in \R^{R \times S}$ & $\partial E / \partial \alpha_{r,s} = \sum_{k=1}^{K} \sum_{c=1}^{C}  \partial E / \partial W_{k,c,r,s}$ \\
  \hline Layer pruning & $\alpha \in \R$ & $\partial E / \partial \alpha = \sum_{k=1}^{K}\sum_{c=1}^{C} \sum_{r=1}^{R} \sum_{s=1}^{S} \partial E / \partial W_{k,c,r,s}$ \\
  \hline
 \end{tabular}
\end{table}

\subsection{Regularization}
\label{sec:regularization}
We use SGD for training and apply momentum and weight decay when updating the dense structure parameters:
\begin{equation}
\label{eq:update_l2}
\Delta \alpha_i(t+1) = \mu \Delta \alpha_i(t) - \eta \frac{\partial E}{\partial \alpha} - \lambda \eta \alpha_i , 
\end{equation}
where $\mu$ is the momentum, $\eta$ is the learning rate and $\lambda$ is the regularization strength.
We use a momentum in order to diminish fluctuations over iterations in parameter changes, which is highly important since we update large structures of a layer.

\begin{figure*}
\centering
\begin{subfigure}{.25\textwidth}
  \centering
  \includegraphics[width=0.9\linewidth]{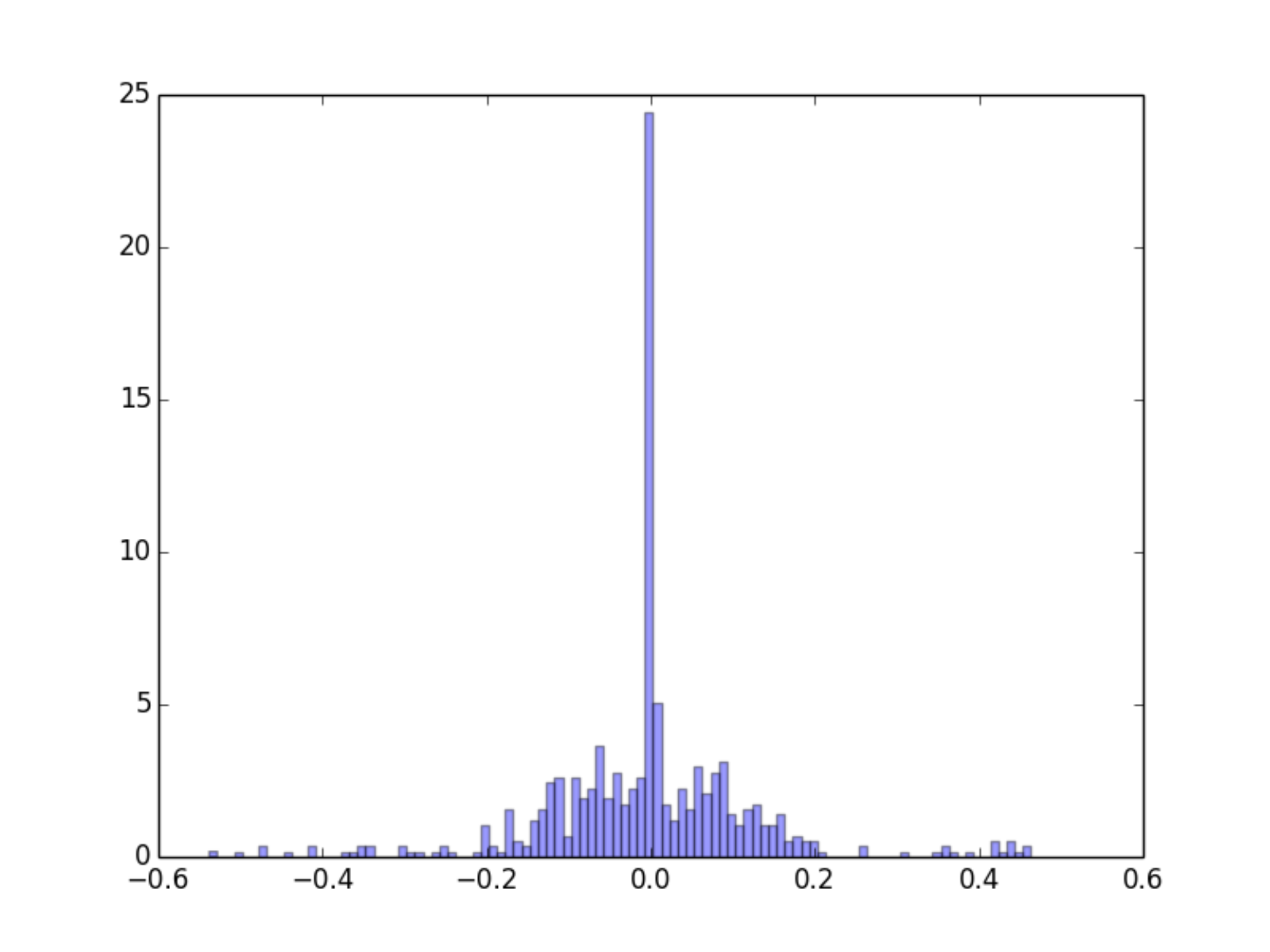}
  \caption{Weight decay; group 0.}
  \label{fig2:sub1}
\end{subfigure}%
\begin{subfigure}{.25\textwidth}
  \centering
  \includegraphics[width=0.9\linewidth]{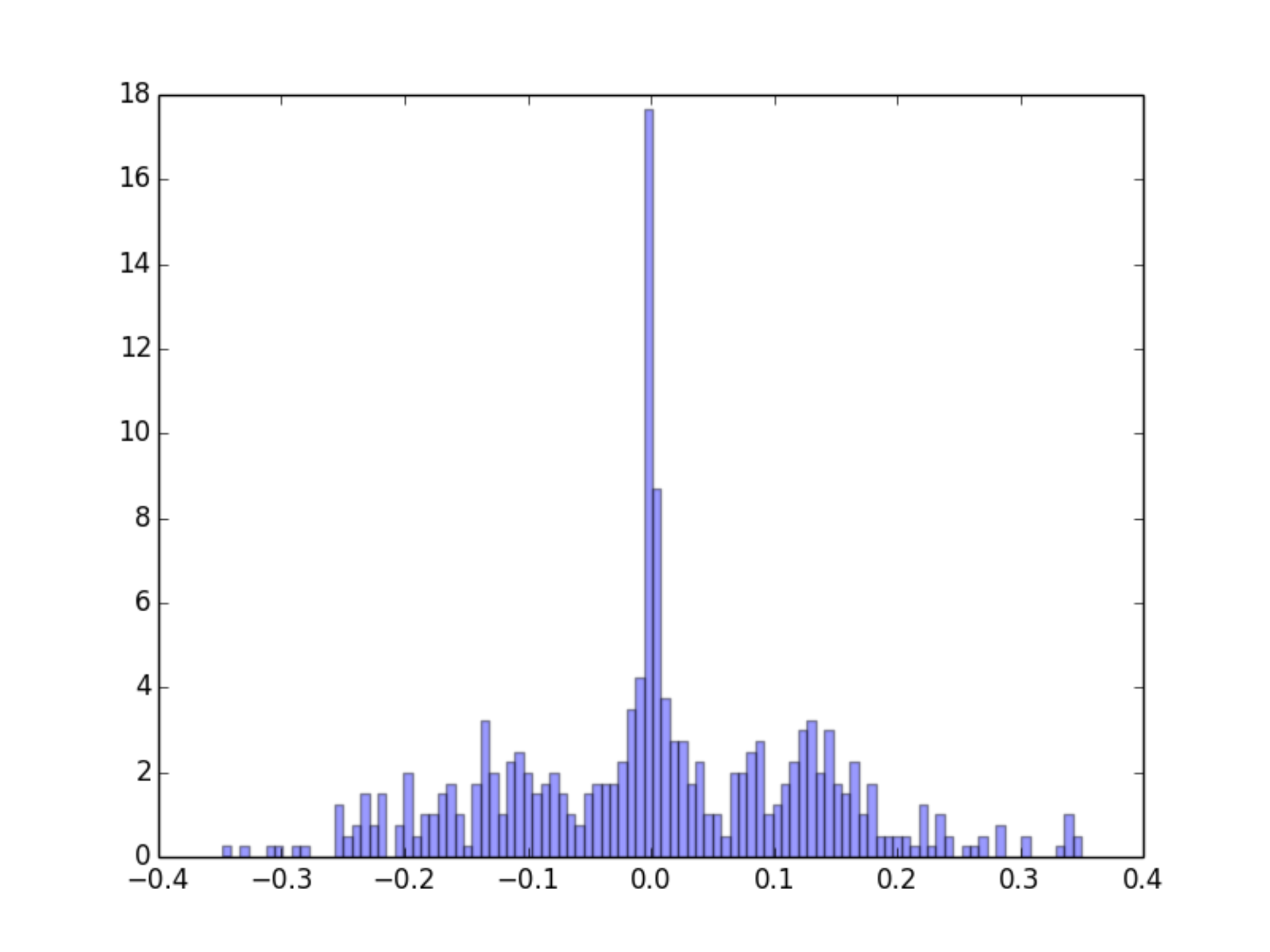}
  \caption{Weight decay; group 1.}
  \label{fig2:sub2}
 \end{subfigure}%
\begin{subfigure}{.25\textwidth}
  \centering
  \includegraphics[width=0.9\linewidth]{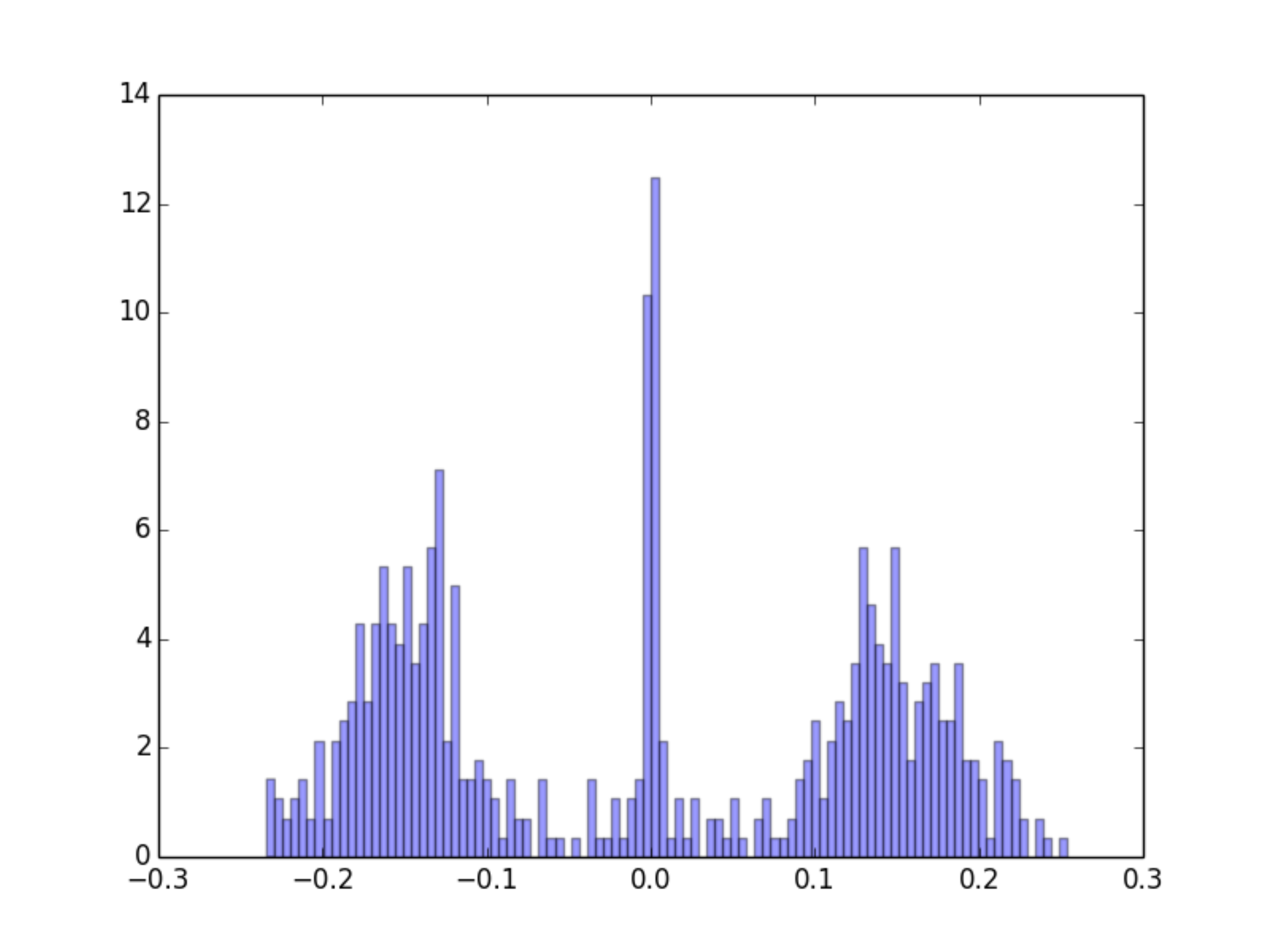}
  \caption{Weight decay; group 2.}
  \label{fig2:sub3}
\end{subfigure}%
\begin{subfigure}{.25\textwidth}
  \centering
  \includegraphics[width=0.9\linewidth]{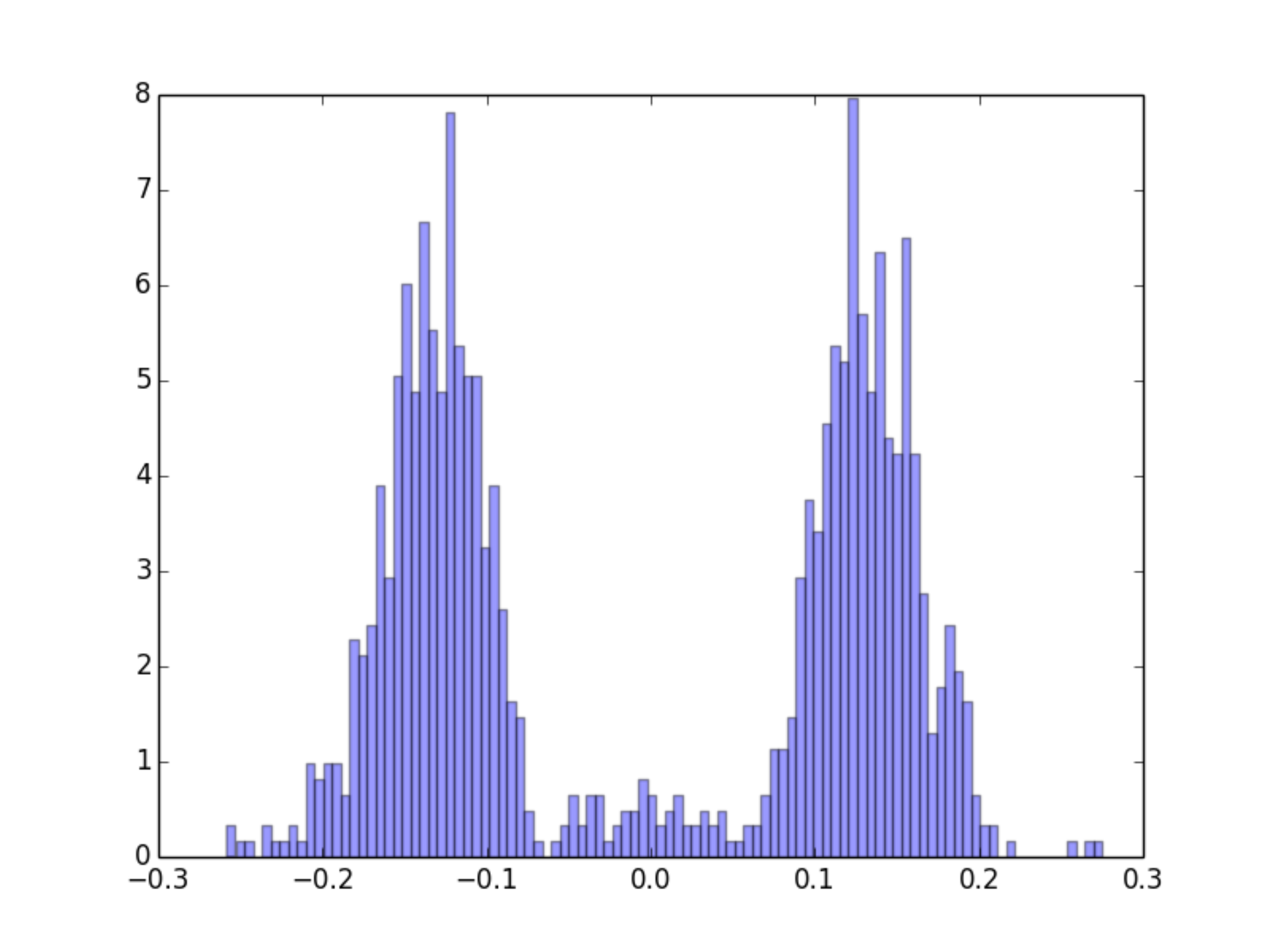}
  \caption{Weight decay; group 3.}
  \label{fig2:sub4}
\end{subfigure}

\begin{subfigure}{.25\textwidth}
  \centering
  \includegraphics[width=0.9\linewidth]{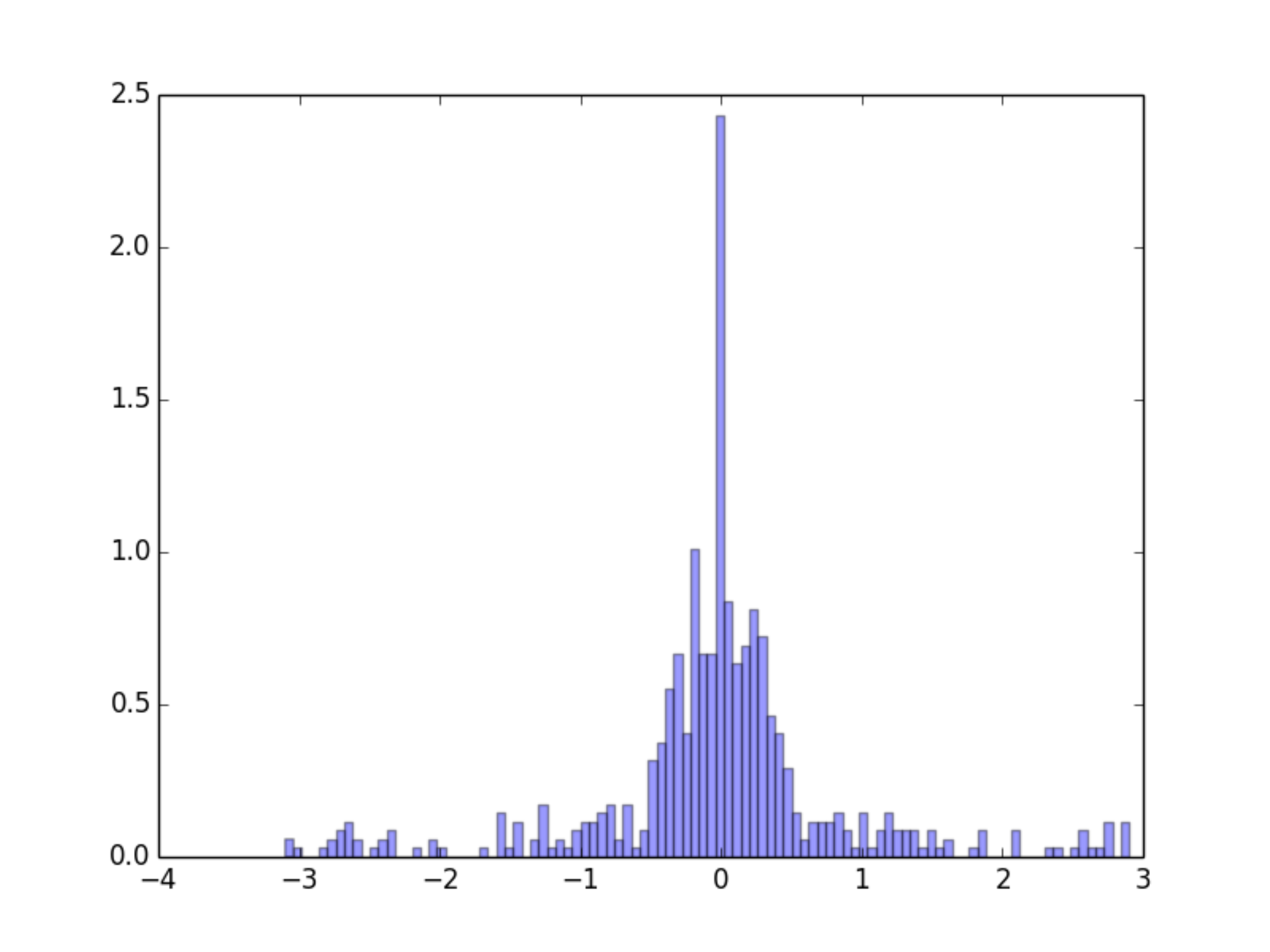}
  \caption{$\ell_1$ reg.; group 0.}
  \label{fig2_2:sub1}
\end{subfigure}%
\begin{subfigure}{.25\textwidth}
  \centering
  \includegraphics[width=0.9\linewidth]{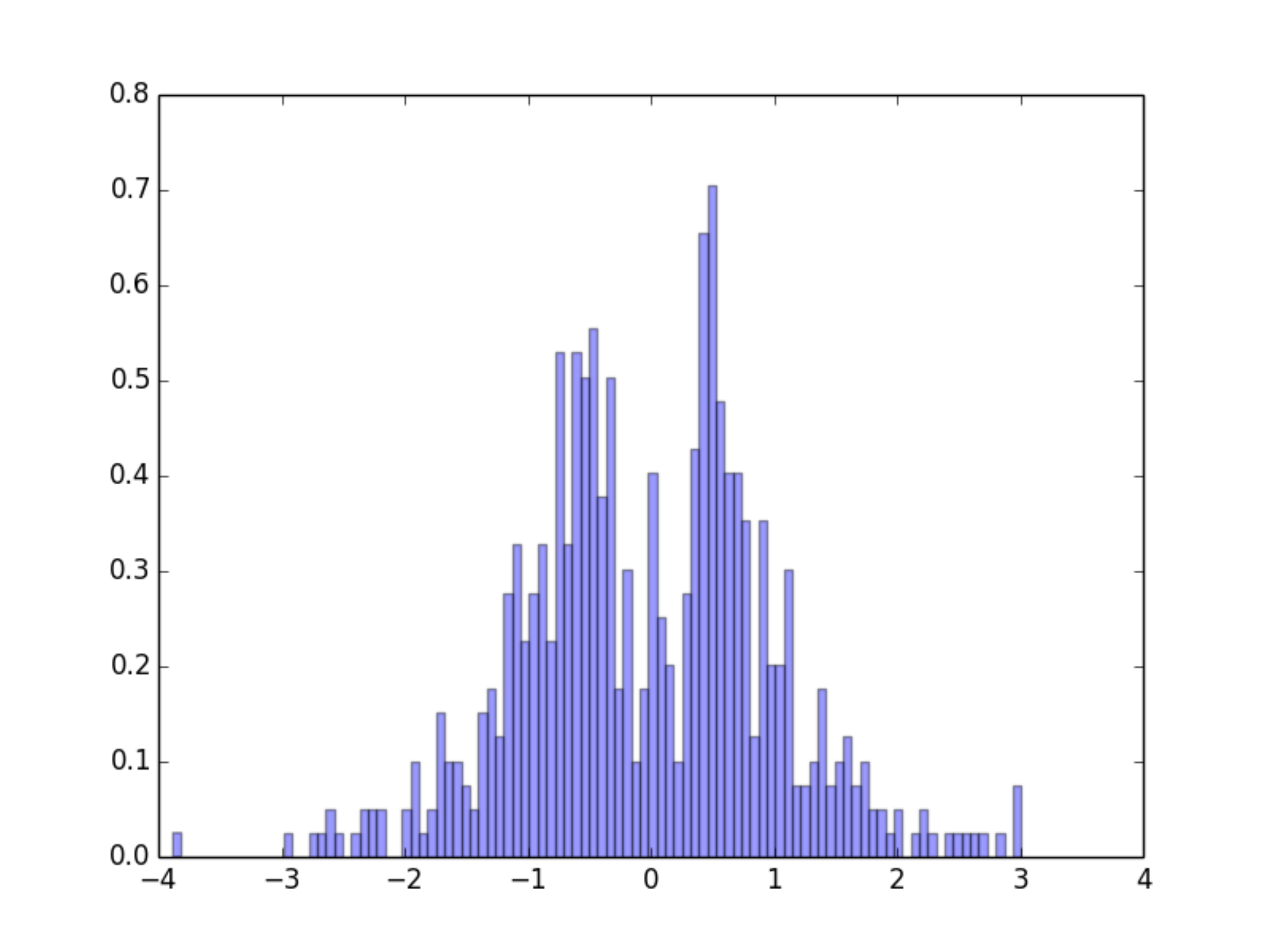}
  \caption{$\ell_1$ reg.; group 1.}
  \label{fig2_2:sub2}
 \end{subfigure}%
\begin{subfigure}{.25\textwidth}
  \centering
  \includegraphics[width=0.9\linewidth]{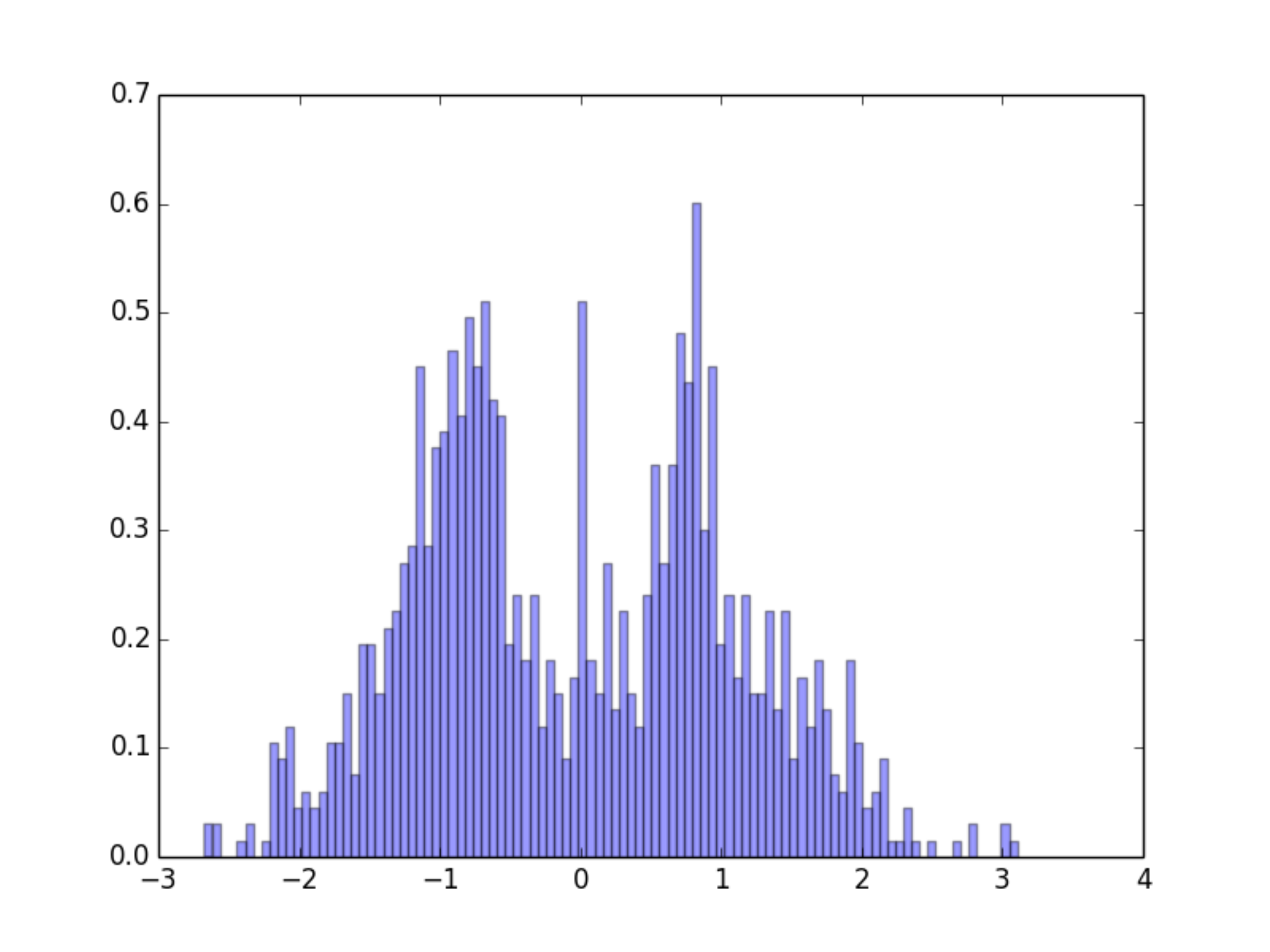}
  \caption{$\ell_1$ reg.; group 2.}
  \label{fig2_2:sub3}
\end{subfigure}%
\begin{subfigure}{.25\textwidth}
  \centering
  \includegraphics[width=0.9\linewidth]{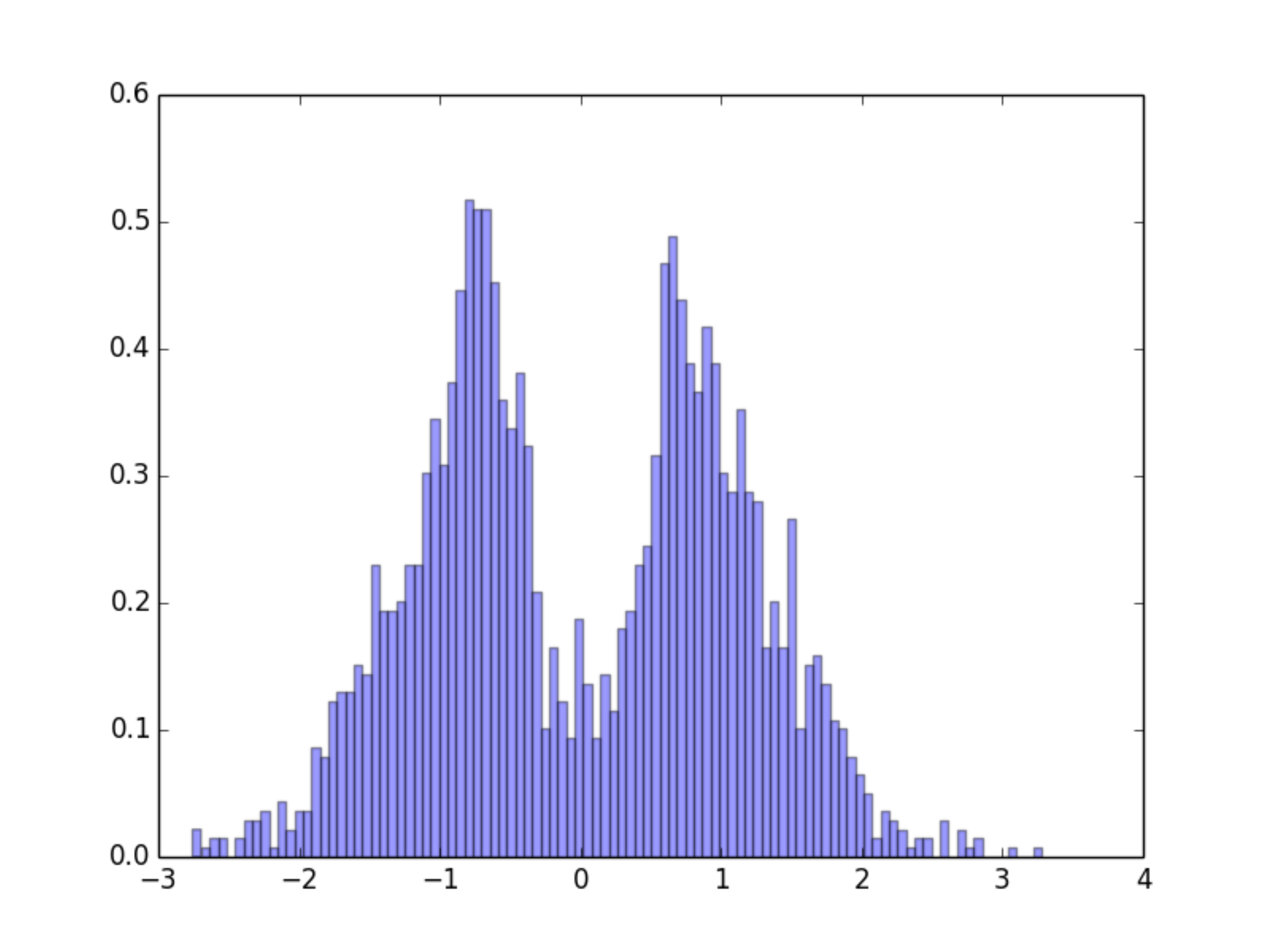}
  \caption{$\ell_1$ reg.; group 3.}
  \label{fig2_2:sub4}
\end{subfigure}
\caption{Different distributions of column-wise structure parameters with weight decay and $\ell_1$ regularization of a fully trained ResNet with 18 layers on ImageNet. The distributions correspond to the first convolution in the first block in the respective group. \textbf{No} pruning was performed ($\epsilon=0$). Note that peaks visually close to zero are not exactly zero.}
\label{fig:distributions1}
\end{figure*}

Regularization not only prevents overfitting, but also decays the dense structure parameters towards zero (but not exactly to zero) and, hence, reduces the pruning error.
Using weight decay for sparsity instead of $\ell_1$ regularization may seem counterintuitive, since $\ell_1$ implicitly decays parameters exactly to zero, however, the update rule between $\ell_1$ regularization and weight decay differs significantly:
the objective function for $\ell_1$ regularization changes to $E_{\ell_1}( \alpha_i)=E(\alpha_i)+\lambda  \abs{\alpha_i}$, while for weight decay it changes to $E_{\ell_2}(\alpha_i)=E(\alpha_i)+ \frac{\lambda}{2} \alpha_i^2$.
Adding the $\ell_1$ penalty results in an SGD update rule as:
\begin{equation}
\label{eq:update_l1}
\Delta \alpha_i(t+1) = \mu \Delta \alpha_i(t) - \eta \frac{\partial E}{\partial \alpha_i} - \lambda \eta \sign(\alpha_i) , 
\end{equation}
while weight decay results in the update rule of Eq. \ref{eq:update_l2}.
$\ell_1$ regularization only considers the direction the parameters are decayed towards and weight decay also takes the magnitude of the parameters into account.
This makes a severe difference in the learning capabilities of SGD based neural networks, that can be best visualized using the distributions of the dense structure parameters $\alpha_i$ (corresponding to different layers) in Fig. \ref{fig:distributions1}.
Parameterizing structures and regularization ultimately shrink the complexity (variance of the layers) of a neural network.
We observe that weight decay without pruning ($\epsilon=0$) produces unimodal, bimodal and trimodal distributions (Fig. \ref{fig2:sub1}-\ref{fig2:sub4}), indicating different complexities, with a clear distinction between important and unimportant dense structure parameters.
In contrast, $\ell_1$ regularization without pruning ($\epsilon=0$) (Fig. \ref{fig2_2:sub1}-Fig. \ref{fig2_2:sub4}) lacks the ability to form this clear distinction.
Second, $\ell_1$ regularized dense structure parameters are roughly one order of magnitude larger than parameters trained with weight decay, making them more sensitive to small noise in the input data.


\section{Experiments}
\label{sec:experiments}
The CIFAR-10 and CIFAR-100 datasets \cite{cifar} consist of colored $32 \times 32$ images, with 50,000 training and 10,000 validation images.
They differ in the number of classes, being 10 respectively 100.
For data augmentation, we subtract the per-pixel mean from the $32 \times 32$ input images, following He et al. \cite{DBLP:journals/corr/HeZRS15}.
The ILSVRC 2012 dataset (ImageNet) \cite{DBLP:journals/corr/RussakovskyDSKSMHKKBBF14} consists of 1.28 million trainings and 50,000 validation images.
We adopt the data preprocessing from \cite{DBLP:journals/corr/HeZRS15,DBLP:journals/corr/HuangLW16a} and we report top-1 and top-5 classification errors on the validation set.

We only use already optimized state-of-the-art networks for our experiments: ResNet \cite{DBLP:journals/corr/HeZRS15} and DenseNet \cite{DBLP:journals/corr/HuangLW16a}.
We use the same networks for CIFAR and ImageNet as described in the original publications.
Both networks apply $1\times1$ convolutions as bottleneck layers before the $3\times3$ convolutions to improve compute and memory efficiency.
DenseNet further improves model compactness by reducing the number of feature maps at transition layers.
If bottleneck and transition compression is used, the models are labeled as \emph{ResNet-B} and \emph{DenseNet-BC}, respectively.
Removing the bottleneck layers in combination with our compression approach has the advantage of reducing both, memory/compute requirements and the depth of the networks.
We apply PSP to all convolutional layers except the sensitive input, output, transition and shortcut layers, which have negligible impact on overall memory and compute costs.

We train all models using SGD and a batch size of 64 (1 GPU) and 256 (8 GPUs) for CIFAR and ImageNet, respectively.
For the CIFAR experiments, we train for 300 epochs and start with a learning rate of 0.1, which is divided by 10 at 50\% and 75\% of the training epochs \cite{DBLP:journals/corr/HuangLW16a}.
For the ImageNet experiments, we train for 110 epochs and start with a learning rate of 0.1, which is divided by 10 at 30, 60, 90 and 100 epochs \cite{DBLP:journals/corr/HeZRS15}.
We use a weight decay of $10^{-4}$ and a momentum of 0.9 for weights and structure parameters throughout this work.
We use the initialization introduced by He et al. \cite{DBLP:journals/corr/HeZR015} for the weights and initialize the structure parameters randomly using a zero-mean Gaussian with standard deviation 0.1.
For the DenseNet experiments, we set the threshold parameter $\epsilon=0.1$ for inducing sparsity (Eq. \ref{eq:threshold}).
For the ResNet experiments, we set the threshold parameter $\epsilon=0.2$, except for the following ablation experiments (Sec. \ref{sec:abl_study}), where we evaluate the sensitivity of the hyperparameter $\epsilon$ and different sparsity constraints.

\subsection{Ablation experiments}
\label{sec:abl_study}
\begin{figure}[t]
\begin{center}
   \includegraphics[width=0.45\linewidth]{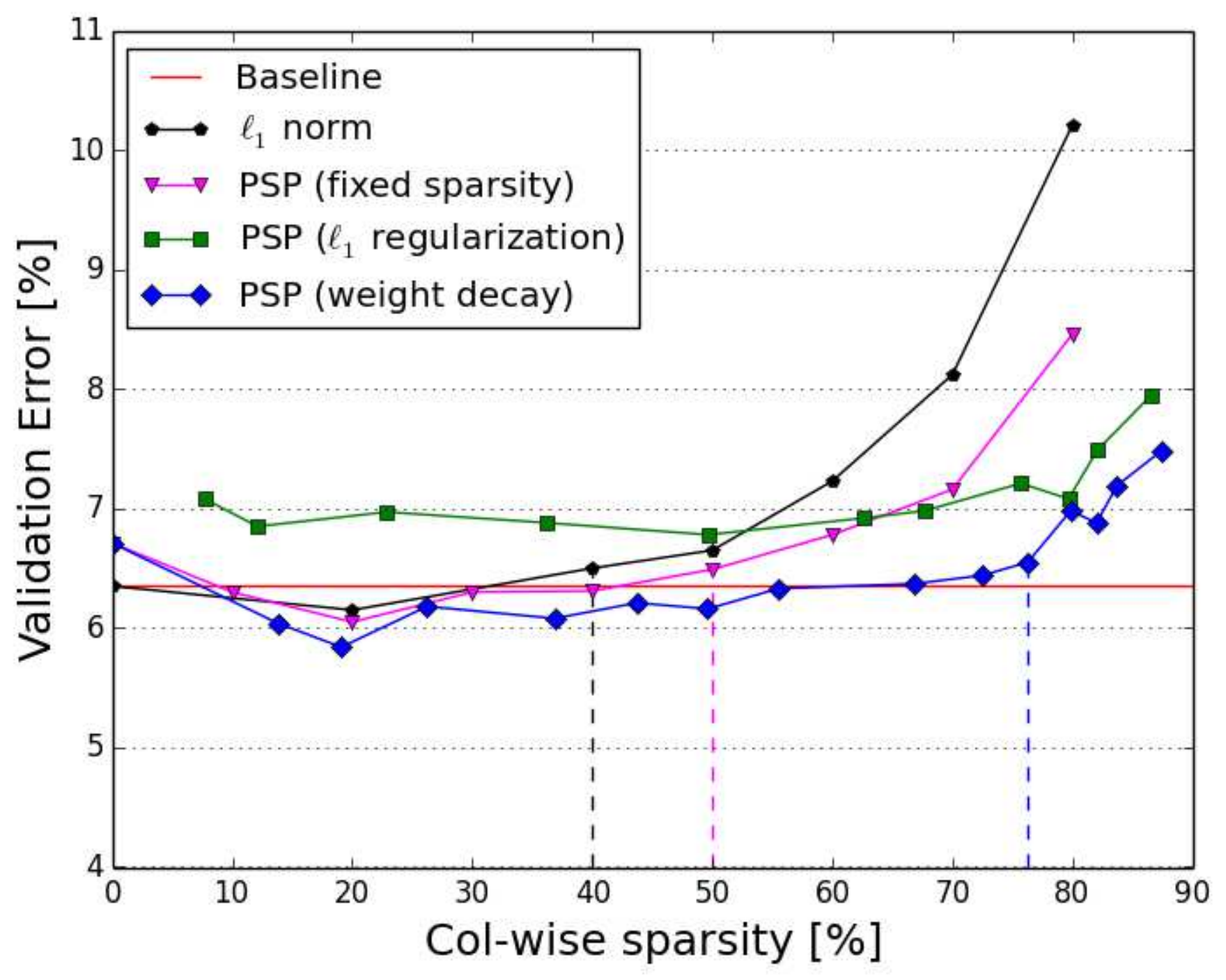}
\end{center}
   \caption{ResNet network with 56 layers on CIFAR10 and column pruning.}
\label{fig:abl_study}
\end{figure}

We start the experiments with an ablation experiment to validate methods and statements made in this work.
This experiment is evaluated on the ResNet architecture, using column pruning, with 56 layers using the CIFAR10 dataset (Fig. \ref{fig:abl_study}).
We report the validation error for varying sparsity constraints, and with the baseline error set to the original unpruned network, with some latitude to filter out fluctuations: $6.35 \% \pm 0.25$.
The dashed vertical lines indicate the maximum amount of sparsity while maintaining the baseline error.
A common way \cite{DBLP:journals/corr/MaoHPLLWD17} to estimate the importance of structures is the $\ell_1$ norm of the targeted structure in a weight tensor $A_{norm} =  \abs{\abs{\text{W}}}_1$, which is followed by pruning the structures with the smallest norm.
We use this rather simple approach as a baseline, denoted as \textit{$\ell_1$ norm}, to show the differences to the proposed parameterized structure pruning.
The parameterization in its most basic form is denoted as \textit{PSP (fixed sparsity)}, where we do not apply regularization ($\lambda=0$ in the SGD update in Eq. \ref{eq:update_l2}) and simply prune the parameters with the lowest magnitude.
As can be seen, this parameterization achieves about 10\% more sparsity compared to the baseline ($\ell_1$ norm) approach, or 1.8\% better accuracy under a sparsity constraint of 80\%.

Furthermore, we observe that regularized dense structure parameters are able to learn a clear distinction between important and unimportant structures (Sec. \ref{sec:regularization}).
Thus, it seems appropriate to use a simple threshold heuristic (Eq. \ref{eq:threshold}) rather than pruning all layers equally (as compared to \textit{PSP (fixed sparsity)}).
We also show the impact of the threshold heuristic in combination with $\ell_1$ regularization (Eq. \ref{eq:update_l1}) and weight decay (Eq. \ref{eq:update_l2}) in Fig. \ref{fig:abl_study}.
These methods are denoted as \textit{PSP ($\ell_1$ regularization)} and \textit{PSP (weight decay)}, respectively.
We vary the regularization strength for $\ell_1$ regularization, since it induces sparsity implicitly, while we vary the threshold parameter for weight decay:
for \textit{PSP ($\ell_1$ regularization)}, we set the threshold $\epsilon=10^{-3}$ and the initial regularization strength $\lambda=10^{-10}$, which is changed by an order of magnitude ($\times10$) to show various sparsity levels.
For \textit{PSP}, we set the regularization strength $\lambda=10^{-4}$ and the initial threshold $\epsilon=0.0$ and increase $\epsilon$ by $2 \cdot 10^{-2}$ for each sparsity level.
Both methods show higher accuracy for high sparsity constraints (sparsity $\ge 80\%$), but only weight decay achieves baseline accuracy.

\subsection{Pruning different structures}
Next, we compare the performance of the different structure granularities using DenseNet on CIFAR10 (Table \ref{tab:granularities}, with 40 layers, a growth rate of $k=12$ and a pruning threshold of $\epsilon=0.1$).
We report the required layers, parameters and Multiply-Accumulate (MAC) operations.
\begin{table*}[t]
\caption{Column-, channel-, shape- and layer-pruning using PSP, validated on DenseNet40 ($k=12$) on the CIFAR10 dataset. \emph{M} and \emph{G} represents $10^6$ and $10^9$, respectively.}
\label{tab:granularities}
\begin{center}
\resizebox{8cm}{!}{%
\begin{tabular}{lr|rrr}
Model	 								& Layers			& Parameters 	&	 MACs		&	Error [\%]  			\\
\hline 
Baseline								& 	40				& 	1.02M			&	0.53G		&	5.80						\\  
Column pruning					& 40					& 	0.22M			&	0.10G		&	5.76						\\
Channel pruning					& 	40				& 	0.35M			&	0.18G		&	5.61						\\  
Shape pruning						& 	40				& 	0.92M			&	0.47G		&	5.40						\\  
Layer pruning						& 	28				& 	0.55M			&	0.28G		&	6.46						\\    
Layer+channel	 pruning		& 	33				& 	0.48M			&	0.24G		&	6.39						\\    
\end{tabular}
}
\end{center}
\end{table*}
While all structure granularities show a good prediction performance, with slight deviations compared to the baseline error, column- and channel-pruning achieve the highest compression ratios.
Shape pruning results in the best accuracy but only at a small compression rate, indicating that a higher pruning threshold is more appropriate.
It is worth noticing that PSP is able to automatically remove structures, which can be seen best when comparing layer pruning and a combination of layer and channel pruning:
layer pruning removes 12 layers from the network but still requires 0.55M parameters and 0.28G MACs, while the combination of layer and channel pruning removes only 7 layers but requires only 0.48M parameters and 0.24G MACs.

\subsection{CIFAR10/100 and ImageNet}
To validate the effectiveness of PSP, we now discuss results from ResNet and DenseNet on CIFAR10/100 and ImageNet.
We use column pruning throughout this section, as it offers the highest compression rates while preserving classification performance.

Table \ref{tab:cifar} reports results for CIFAR10/100.
As can be seen, PSP maintains classification performance for a variety of networks and datasets.
This is due to the ability of self-adapting the pruned structures during training, which can be best seen when changing the network topology or dataset:
for instance, when we use the same models on CIFAR10 and the more complex CIFAR100 task, we can see that PSP is able to automatically adapt as it removes less structure from the network trained on CIFAR100.
Furthermore, if we increase the number of layers by $2.5\times$ from 40 to 100, we also increase the over-parameterization of the network and PSP automatically removes $2.4\times$ more structure.
\begin{table*}[t]
\caption{ResNet and DenseNet on CIFAR10/100 using column pruning. \emph{M} and \emph{G} represents $10^6$ and $10^9$, respectively.}
\label{tab:cifar}
\begin{center}
\resizebox{11cm}{!}{%
\begin{tabular}{lr|rrr|rrr}
			 								& 				& \multicolumn{3}{c|}{CIFAR10} 				  	&		\multicolumn{3}{c}{CIFAR100}  				\\
	 										& 				&  				&	 				&	Error 	&			 	&	 				&	Error   	\\
Model	 								& Layer			& Parameter 	&	 MACs		&	[\%]  	&	Parameter 	&	 MACs			&	[\%]  		\\
\hline 
NASNet-B (4 @ 1152)	\cite{DBLP:journals/corr/ZophVSL17}		
											& 	--					& 	2.60M			&	--				&	3.73				&	--					&	--				&	--								\\  
\hline 
ResNet									& 	56				& 	0.85M			& 0.13G		&	6.35				&	0.86M			&	0.13G		&	27.79		\\  
ResNet-PSP							& 	56				& 	0.21M			& 0.03G		&	6.55				&	0.45M			&	0.07G		&	27.15		\\  
\hline 
DenseNet	($k=12$)				& 	40				& 	1.02M			&	0.27G		&	5.80				&	1.06M			&	0.27G		&	26.43						\\  
DenseNet-PSP	($k=12$)		& 	40				& 	0.22M			&	0.05G		&	5.76				&	0.37M			&	0.08G		&	26.30					\\  
\hline 
DenseNet	($k=12$)				& 	100				& 	6.98M			&	1.77G		&	4.67				&	7.09M			& 1.77G		&	22.83						\\  
DenseNet-PSP ($k=12$)		& 	100				& 	0.99M			&	0.22G		&	4.87				&	1.17M			&	0.24G		&	23.42						\\  
\end{tabular}
}
\end{center}
\end{table*}

The same tendencies can be observed on the large-scale ImageNet task as shown in Table \ref{tab:imagenet}; when applying PSP, classification accuracy can be maintained (with some negligible degradation) and a considerable amount of structure can be removed from the networks (e.g. $2.6\times$ from ResNet18 or $1.8\times$ from DenseNet121).
Furthermore, PSP obliterates the need for $1\times1$ bottleneck layers, effectively reducing network depth and MACs.
For instance, removing the bottleneck layers from the DenseNet121 network in combination with PSP removes $2.6\times$ parameters, $4.9\times$ MACs and $1.9\times$ layers, while only sacrificing $2.28\%$ top-5 accuracy.
\begin{table*}[t]
\caption{ResNet and DenseNet on ImageNet using column pruning.}
\label{tab:imagenet}
\begin{center}
\resizebox{9cm}{!}{%
\begin{tabular}{lr|rrrr}
Model	 					&	 Layer		&	Parameters  	&	MACs		& Top-1 [\%]	& Top-5 [\%]
\\ \hline 
MobileNetV2 (1.4) \cite{DBLP:journals/corr/abs-1801-04381}
								& --				& 6.9M				& 0.59G		& 25.30			& --							\\
NASNet-A (4 \@ 1056) \cite{DBLP:journals/corr/ZophVSL17}
								&					&	5.3M				&	0.56G			& 	26.00	& 8.40				
\\ \hline 
ResNet-B					&	18			&	11.85M		&	1.82G		& 	29.60			& 	10.52				\\  
ResNet-B-PSP			&	18			&	5.65M			&	0.82G		& 	30.37			& 	11.10						\\ 
 \hline 
ResNet-B					&	50			&	25.61M		& 4.09G		& 	23.68			& 	6.85							\\
ResNet-B-PSP			&	50			&	15.08M		&	2.26G		& 	24.07			& 	6.69						\\
\hline 
DenseNet-BC			&	121			&	7.91M			&	2.84G		& 	25.65			& 	8.34							\\
DenseNet-BC-PSP		&	121			&	4.38M			&	1.38G		& 	25.95			& 	8.29							\\	
\hline 
DenseNet-C				&	63			&	10.80M		&	3.05G		& 	28.87			& 	10.02						\\
DenseNet-C-PSP		&	63			&	3.03M			&	0.58G		& 	29.66			& 	10.62						\\
\hline 
DenseNet-C				&	87			&	23.66M		&	5.23G		& 	26.31			& 	8.55							\\
DenseNet-C-PSP		&	87			&	4.87M			&	0.82G		& 	27.46			& 	9.15							\\
\end{tabular}
}
\end{center}
\end{table*}

We also report some results of recently proposed network reduction methods that achieved notable performance on the used datasets (in terms of accuracy, memory and compute requirements):
MobileNetV2 \cite{DBLP:journals/corr/abs-1801-04381} is an optimized CNN network for mobile platforms, which uses, among other optimizations, the popular lightweight depthwise convolutions.
NASNet \cite{DBLP:journals/corr/ZophVSL17} is a Neural Network Search (NAS) algorithm that searches for the best neural network architecture.
PSP outperforms the efficiency of these methods, using standard networks and requiring only a fraction of the training time of NAS.

\subsection{Comparison to related methods}
We report a profound comparison to related structured pruning methods (see Sec. \ref{sec:related_work} for details) in Table \ref{tab:comparison}.
As reported metrics and baseline accuracy vary significantly in the corresponding publications, to show a fair comparison, we only report the improvement factor and the accuracy gap over the baseline network, where $+$ represents accuracy degradation and $-$ accuracy improvement.
\begin{table*}[t]
\caption{Comparison to related structured pruning methods on a variety of networks and datasets.}
\label{tab:comparison}
\begin{center}
\resizebox{10cm}{!}{%
\begin{tabular}{lcccccccc}
			 							& ThiNet
			 							&  CP 		
			 							&	 DCP	
			 							&	Slimming 
			 							&	SPP
			 							&	LCP	
			 							&	SFP	
			 							&	PSP  \\
			 							& \cite{DBLP:journals/corr/abs-1810-11809} 	
			 							& \cite{DBLP:journals/corr/HeZS17}
			 							&	 \cite{DBLP:journals/corr/abs-1810-11809}	
			 							&	\cite{DBLP:journals/corr/abs-1708-06519}
			 							&	\cite{DBLP:journals/corr/abs-1709-06994}
			 							&	\cite{DBLP:journals/corr/abs-1810-00518}	
			 							&	\cite{DBLP:journals/corr/abs-1808-06866}	
			 							&  (ours)
\\ \hline 
\multicolumn{9}{c}{ResNet-56 on CIFAR10: error=6.35\%}								
\\ \hline 
Parameters						&	1.97x		&  --			&	1.97x 	&	--			&	--			&	--			&	--			&	3.86x			\\
FLOPs								&	1.99x		&  2.00x	&	1.99x	&	--			&	--			&	2.00x	&	2.11x	&	4.17x			\\
Error gap							&	+0.82		&  +1.00	&	+0.31	&	--			&	--			&	+0.77	&	+0.24	&	+0.20
\\ \hline 
\multicolumn{9}{c}{DenseNet-40 on CIFAR10: error=5.80\%}
\\ \hline 
Parameters						&	--				&  --			&	--		 	&	2.87x	&	--			&	--			&	--			&	4.64x			\\
FLOPs								&	--				& 	--			&	--			&	2.22x	&	--			&	--			&	--			&	5.30x			\\
Error gap	 						&	--				&  --			&	--			&	-0.46	&	--			&	--			&	--			&	-0.03			\\
 \hline 
\multicolumn{9}{c}{ResNet-18 on ImageNet: top1 error=29.60\%, top5 error=10.52\%}
\\ \hline
Parameters						&	--				&  --			&	2.00x 	&	--			&	--			&	--			&	--			&	2.10x			\\
FLOPs								&	--				&  --			&	2.00x	&	--			&	--			&	--			&	1.72x	&	2.22x			\\
Top1 error gap					&	--				&  --			&	+2.29	&	--			&	--			&	--			&	+3.18	&	+0.77			\\
Top5 error gap					&	--				&  --			&	+1.38	&	--			&	--			&	--			&	+1.85	&	+0.58			\\
 \hline 
\multicolumn{9}{c}{ResNet-B-50 on ImageNet: top1 error=23.68\%, top5 error=6.85\%}
\\ \hline
Parameters						&	2.06x		&  --			&	2.06x 	&	--			&	--			&	--			&	--			&	1.70x				\\
FLOPs								&	2.25x		&  2.00x	&	2.25x	&	--			&	2.00x	&	2.00x	&	1.72x	&	1.81x				\\
Top1 error gap					&	+1.87		&  --			&	+1.06	&	--			&	--			&	+0.96	&	+1.54	&	+0.39				\\
Top5 error gap					&	+1.12		&  +1.40	&	+0.61	&	--			&	+0.8		&	+0.42	&	+0.81	&	+0.16				\\
\end{tabular}
}
\end{center}
\end{table*}
As can be seen, PSP outperforms other pruning methods substantially in memory, compute requirements, and accuracy.
Due to the self-adapting pruning method, PSP achieves less compression on ResNet-B-50 on ImageNet, however, it achieves the best accuracy and is inline with overarching goals.


\section{Conclusion}
\label{sec:conclusion}
We have presented PSP, a novel approach for compressing DNNs through structured pruning, which reduces memory and compute requirements while creating a form of sparsity that is inline with massively parallel processors.
Our approach exhibits parameterization of arbitrary structures (e.g. channels or layers) in a weight tensor and uses weight decay to force certain structures towards zero, while clearly discriminating between important and unimportant structures.
Combined with threshold-based magnitude pruning and backward approximation, we can remove a large amount of structure while maintaining prediction performance.
Experiments using state-of-the-art DNN architectures on real-world tasks show the effectiveness of our approach in comparison to a variety of related methods.
As a result, the gap between DNN-based application demand and capabilities of resource-constrained devices is reduced, while this method is applicable to a wide range of processors.

\printbibliography

\end{document}